\documentclass[conference]{IEEEtran}
\usepackage[T1]{fontenc}
\usepackage{graphicx}
\usepackage{booktabs}
\usepackage{amsmath}
\usepackage[numbers,sort&compress]{natbib}
\usepackage[colorlinks=true,linkcolor=blue,citecolor=blue,urlcolor=blue]{hyperref}
\usepackage{microtype}
\usepackage{xcolor}

\title{GIFT: Glove-Inferred Force Transfer\\
\large Force-Aware Human-to-Robot Skill Transfer from a Wearable Sensing Glove\\
to a Robot Hand Without Tactile Sensors}
\author{\IEEEauthorblockN{Tzach Sarusi\thanks{Project page, videos and the evaluation logs: \url{https://tzahsarusi.github.io}.}}
\IEEEauthorblockA{tzahsarusi@gmail.com}}

\IEEEoverridecommandlockouts
\begin{document}
\maketitle

\begin{abstract}
Human-to-robot skill transfer from sensing gloves has so far relied on shared
hardware: the same tactile glove worn by the demonstrator and the robot, or a
learned alignment between two tactile sensors. We present GIFT (Glove-Inferred
Force Transfer), a pipeline in which the interface between human and robot is a
physical unit rather than a shared sensor: fingertip force is measured in
newtons on the human side and estimated in newtons on the robot side. A
wearable glove records finger flexion, calibrated fingertip force, and wrist
orientation, while a head-mounted camera records the demonstration; no robot is
present during data collection. At deployment, the robot estimates force from
actuator-current residuals relative to a free-space baseline, through a
calibrated mapping to newtons, so any position-controlled hand that reports
motor current can serve as the deployment platform. The robot hand carries no
tactile sensors. The policy uses a glove-space state representation and
predicts finger-position targets; the robot enters only through two calibrated
adapters, a retargeting decoder and a force estimator. We evaluate GIFT on a
cup grasp-and-hold task with two action-chunking policies trained on the same
demonstrations, with fingertip-force inputs retained in one and zeroed in the
other. In a 50-rollout evaluation with sample size and metrics fixed before
scoring, both policies succeeded in all 25 rollouts. The median of the
per-rollout hold-phase grip-force estimates was 53\% lower with force inputs:
1.20\,N versus 2.55\,N (one-sided Mann--Whitney U, $p<0.0001$). In an
observation ablation, a vision-only policy achieved 0/15 grasps, policies given
hand-command state acquired the grasp, and the force inputs determined how
hard the policy held. A force channel measured on the human hand thus
transfers to a robot hand with no tactile hardware, through a retargeting map
from five glove channels to seven robot actuators, with no sensor shared
between the two.
\end{abstract}


\section{Introduction}

\begin{figure}[t]
  \centering
  \includegraphics[width=\linewidth]{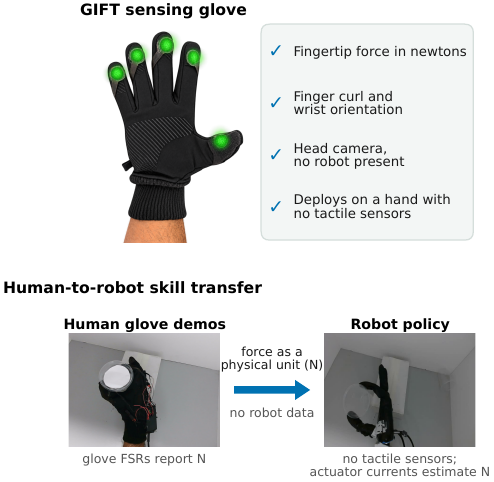}
  \caption{\textbf{GIFT.} Top: the sensing glove (fingertip sensors
  highlighted) captures fingertip force in newtons, finger curl and wrist
  orientation, alongside a head camera; no robot is present during collection. Bottom: the policy trained on these
  demonstrations is deployed on a robot hand with no tactile sensors, whose
  force observation is estimated in newtons from its own actuator currents. The
  human and the robot share a physical unit, not a sensor.}
  \label{fig:teaser}
\end{figure}

Contact-rich manipulation is regulated by touch. A person holding a thin plastic
cup applies just enough grip not to drop it and little enough not to crush it,
and does so without looking, because the controlling signal is force rather
than appearance. Imitation-learning pipelines built on cameras alone must infer
this hidden variable from pixels, and on tasks where the visual scene barely
changes with grip force, they cannot.

The value of tactile signals in human demonstrations is now well established:
tactile gloves transfer human demonstrations to robots~\cite{yin2025osmo,
adeniji2025feel}, and cross-sensor alignment transfers tactile policies between
sensing hardware~\cite{wi2026tactalign}. These pipelines close the
human-to-robot gap by sharing hardware or by learning to bridge it. OSMO puts
the same tactile glove on the demonstrator and the robot, so that the glove
itself is the interface~\cite{yin2025osmo}; TactAlign learns an alignment
between the human's and the robot's tactile sensors~\cite{wi2026tactalign}. In
both cases the robot carries tactile sensing at deployment.

GIFT takes a different route: the interface between human and robot is a
physical unit. Force is recorded in newtons on the glove and estimated in
newtons on the robot, so the two sides never share a sensor, and the robot
needs none. This rests on two design decisions. First, force is captured as a
physical quantity: the glove's fingertip force-sensitive resistors (FSRs) are
calibrated to newtons against a reference scale, so the recorded channel is not
tied to any particular sensor. Second, force at deployment is \emph{estimated}
rather than sensed: following the residual principle of force estimation from
actuation~\cite{oh2026factr2}, the robot hand's actuator currents are compared
against a free-space baseline and mapped into the same newton range the policy
saw in training. The policy itself observes and acts in glove space; the robot
appears only through two calibrated adapters, a retargeting decoder from glove
space to actuator commands and the force estimator. Nothing in the pipeline is
specific to the hand used here; any position-controlled hand that reports motor
current provides the deployment-side signal. The capture rig is a glove with
five flex sensors, five fingertip FSRs, and a wrist IMU, read by an ESP32-S3
microcontroller, together with a head-mounted camera. No robot is present
during data collection (Fig.~\ref{fig:teaser}).

We evaluate GIFT with two action-chunking policies~\cite{zhao2023learning}
trained on the same demonstrations of a cup grasp-and-hold task, with identical
architecture, hyperparameters, and seed, differing in exactly one thing: the
fingertip-force inputs are retained in one policy and zeroed in the other. The
policies are deployed on a seven-actuator tendon-driven hand with no tactile
sensing. Both policies succeeded in all 25 of their rollouts, but the policy
with force inputs held the cup at less than half the estimated grip force.
A third policy trained on vision alone never acquires the grasp, which places
the contribution of each observation channel on a single ladder.

\textbf{Contributions.}
\begin{enumerate}
  \item \textbf{GIFT, an end-to-end pipeline for human-to-robot skill
        transfer of force without tactile hardware on the robot:} a calibrated
        sensing glove and head camera for capture, a policy with a glove-space
        state representation that predicts finger-position targets,
        spline-based retargeting from glove space to hand actuators at runtime,
        and current-residual force estimation at deployment. The pipeline requires no robot during capture and no
        tactile sensor at deployment, and the human and robot share no sensor.
  \item \textbf{A 50-rollout result on real hardware:} at saturated task
        success (25/25 for both policies), the policy with force inputs held at
        \textbf{53\% lower estimated grip force} (median of per-rollout
        hold-phase medians, 1.20\,N versus 2.55\,N, one-sided Mann--Whitney U,
        $p<0.0001$).
  \item \textbf{An observation ablation} isolating the channels: a vision-only
        policy fails to grasp in 0/15 rollouts against 3/3 for in-session
        controls (Fisher exact $p\approx0.0012$); hand-command state acquires
        the grasp; the fingertip-force inputs determine how hard the policy
        holds.
\end{enumerate}

\section{Related Work}

\textbf{Tactile gloves for human-to-robot skill transfer.}
OSMO~\cite{yin2025osmo} is the closest antecedent: an open-source magnetic
tactile glove worn by both the human demonstrator and the robot hand, so that
the glove itself is the shared interface and the visual and tactile embodiment
gaps are minimized. Feel the Force~\cite{adeniji2025feel} collects contact
demonstrations with an AnySkin-augmented glove~\cite{bhirangi2024anyskin} and
reports higher success than policies trained from teleoperation. GIFT differs
in two respects: the capture hardware is piezoresistive (FSRs and flex sensors
rather than magnetic skins), and the robot carries no tactile sensor at all,
so the deployment-side signal is estimated from actuation rather than sensed.
The human and the robot share a unit, not a sensor.

\textbf{Robot-free demonstration capture.}
Handheld and wearable interfaces decouple demonstration collection from robot
access: UMI's handheld gripper~\cite{chi2024universal}, exoskeleton and wearable
dexterous-hand interfaces~\cite{xu2025dexumi, xu2026realdexumi, koh2026dexmouse},
and in-the-wild egocentric pipelines~\cite{tao2025dexwild, zheng2026egoscale}
all scale collection by removing the robot from the loop. These systems
predominantly capture kinematics and vision. GIFT adds calibrated fingertip
force as a first-class captured signal, which our ablation shows to be the
channel that determines grip behavior.

\textbf{Tactile sensing and representations for manipulation policies.}
Dense tactile arrays improve fine-grained manipulation
policies~\cite{huang2024vitac, huang2026flexitac}; vision-language-action
models augmented with tactile inputs gain force-aware
behavior~\cite{huang2025tactilevla}; and representation learning transfers
across heterogeneous tactile sensors~\cite{zhao2024transferable,
higuera2024sparsh}. A complementary line removes deployment-time tactile
hardware by \emph{imagining} tactile signals from
vision~\cite{zhang2026imagining}. GIFT shares the goal of tactile-free
deployment but takes the actuation route: the hand's own motor currents,
baselined and calibrated, stand in for the missing
sensor~\cite{liu2025factr, oh2026factr2}.

\section{The GIFT Pipeline}

GIFT is built around four design principles. (1)~\emph{Force is a physical
unit}: the captured channel is expressed in newtons, so any instrument that
produces newtons can stand in for the glove at deployment. (2)~\emph{No robot
during capture}: demonstrations are recorded on the human hand with a head
camera, so collection scales with people rather than with robots. (3)~\emph{No
tactile sensor at deployment}: the robot's own actuator currents, baselined in
free space, provide the force observation. (4)~\emph{One instrument scores both
policies}: the with-force and without-force policies are compared with the
same deployment-side estimate, so the force channel is the only difference
between them.

The pipeline has four stages: capture on the human with a sensing glove and
head camera (\S\ref{sec:glove}, \S\ref{sec:capture}), retargeting from glove
space to robot actuators (\S\ref{sec:retarget}), policy training on the
demonstrations (\S\ref{sec:policy}), and deployment with force estimated from
actuator current (\S\ref{sec:force}).

\begin{figure*}[t]
  \centering
  \includegraphics[width=\textwidth]{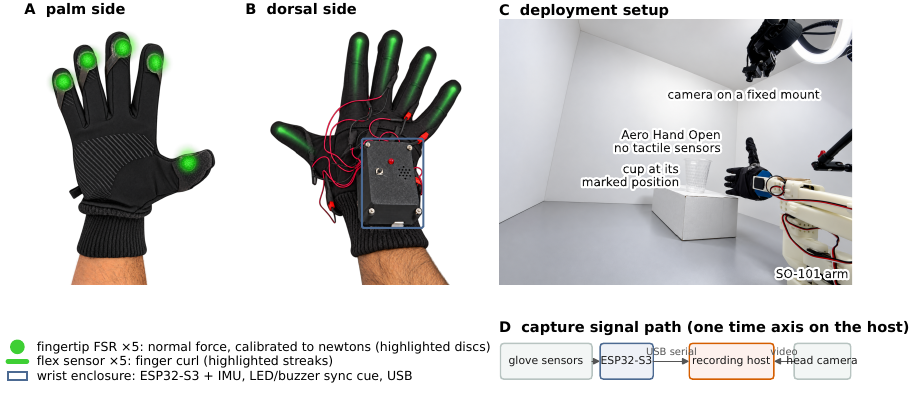}
  \caption{\textbf{Hardware.} (A, B) The sensing glove, palm and dorsal sides,
  with sensor locations highlighted: five fingertip force-sensitive resistors
  (discs) and five flex sensors along the fingers (streaks); the wrist
  enclosure holds the ESP32-S3, the IMU and the LED/buzzer start cue. (C)
  Deployment setup (illustration): the TetherIA Aero Hand Open in a plain
  nitrile glove with no sensors, mounted on an SO-101 arm, the cup at its
  marked position, and the camera on a fixed mount aligned to the
  demonstrator's viewpoint. (D) Capture signal path: all glove channels and the
  camera share one time axis on the recording host.}
  \label{fig:hardware}
\end{figure*}

\subsection{Sensing glove}
\label{sec:glove}
The glove mounts three sensor groups (Fig.~\ref{fig:hardware}). Five flex
sensors, one per finger along the dorsal side, measure finger curl. Five
force-sensitive resistors at the fingertips measure normal contact force. A
WitMotion IMU at the wrist provides wrist orientation (roll, pitch, yaw). All
channels are read by a Seeed XIAO ESP32-S3 microcontroller and streamed over
USB serial to the recording host. Fingertip force is calibrated to newtons
against a digital reference scale with a fitted curve
(Fig.~\ref{fig:instruments}A), so that the recorded force channel is a
physical quantity rather than a raw sensor reading. This is the property that
lets the deployment side substitute a different instrument (\S\ref{sec:force}).

\subsection{Demonstration capture}
\label{sec:capture}
A head-mounted camera records the egocentric view. The demonstrator wears the
glove and performs the task with their own hand; no robot is present or
required. All streams share a single time axis on the recording host, and an
audible start cue captured in both the sensor log and the video verifies
alignment. The demonstration corpus for this work is 73 sessions and
38{,}150 frames of a constant-force cup grasp-and-hold task.

\subsection{Retargeting from glove space to robot actuators}
\label{sec:retarget}
The target platform is a TetherIA Aero Hand Open (seven actuators, sixteen
joints). The policy acts in glove space: its actions are five finger targets in
the glove's flex coordinates, and a per-finger mapping decodes them into
actuator commands at runtime, so retargeting is an adapter around the policy
rather than a transformation of the training data. The thumb, whose single flex
channel drives three joints, uses monotone PCHIP splines fitted through
captured calibration anchors (open pose, intermediate apertures, and the task
hold pose); thumb opposition rides the spline, so as the hand closes the thumb
sweeps into opposition as the demonstrator's did. The hold-pose anchor is
pinned to the median demonstrated hold command, so the policy's hold action
decodes to the demonstrated grip geometry.

\begin{figure*}[t]
  \centering
  \includegraphics[width=\textwidth]{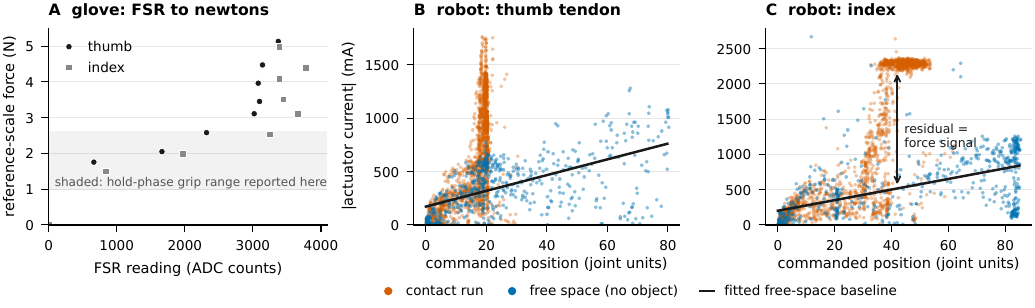}
  \caption{\textbf{Two instruments, one unit.} (A) Glove side: fingertip FSR
  readings against a digital reference scale; the shaded band is the hold-phase
  grip range reported in \S\ref{sec:results}. (B, C) Robot side: actuator
  current against commanded position for the thumb-tendon and index actuators,
  from the free-space log (no object) with its fitted linear baseline, and from
  a contact log. At the grasp position the contact current rises vertically
  above the baseline; that residual, mapped to newtons, is the policy's
  deployment-time force observation.}
  \label{fig:instruments}
\end{figure*}

\subsection{Policy}
\label{sec:policy}
We train action-chunking transformer (ACT) policies~\cite{zhao2023learning} with
LeRobot~\cite{cadene2026lerobot}. The observation is the egocentric image plus a
13-dimensional state vector: five flex, five fingertip forces, and three IMU
angles. The action is the five finger targets in glove flex coordinates
(\S\ref{sec:retarget}). To isolate the force channel we train two policies on
the same corpus. The \textbf{with-force} policy sees all 13 state dimensions;
the \textbf{without-force} policy has the five force dimensions zeroed at
training time. Architecture, preprocessing (a uniform FSR baseline shift and
pinch-symmetry imputation), image augmentation, 20k training steps, batch size,
and seed are identical. At deployment both policies receive identical inputs
except the force channels (robot-side force estimates versus zeros, exactly
as trained).

\subsection{Deployment-time force from actuator current}
\label{sec:force}
The robot hand has no tactile sensors. Following the residual principle of
force estimation from actuation~\cite{oh2026factr2}, we log a free-space
baseline of actuator current versus hand position (no object) and, at
deployment, compute per-actuator residuals
\begin{equation}
i_{\mathrm{contact}} = i_{\mathrm{measured}} - \hat{i}_{\mathrm{free}}(q).
\end{equation}
A tendon-driven hand performing a static hold permits this position-indexed
baseline where a full arm would require a learned dynamics model
(Fig.~\ref{fig:instruments}B,\,C). Residuals map to newtons through a fitted
calibration and are normalized into the same range as the training-set glove
FSRs, keeping the policy's force observation in distribution. The policy's
state therefore stays in glove space at deployment as well: the flex channels
are the policy's own commanded glove-space pose fed back as state (under
position control the hand tracks the command until contact), the IMU channels
are held at values representative of the demonstrations since the wrist
trajectory is fixed, and the force channels carry the current-based estimate in
newtons. The robot enters the loop only through the retargeting decoder and
this estimator. For reporting, force values in Fig.~\ref{fig:chart} and
Table~\ref{tab:ablation} are recomputed from the logged raw signal using a
digital-scale calibration of the robot fingertips; this reporting calibration
was declared before scoring, and runtime behavior used the frozen deployment
estimator throughout.

\section{Experiments}

\subsection{Setup}
\textbf{Platform.} The Aero Hand is mounted on an SO-101 arm executing a
pre-programmed wrist-only trajectory (approach the cup, grasp window, lift,
settle). The lift is triggered when the deployment force estimate reports a
sustained grip ($>0.5$\,N for 1\,s) or after an 18\,s timeout, with the same
rule for both policies. The policy controls the hand only, under position
control with identical torque caps for both policies; the action chunk length
matches training ($n_{\mathrm{action\_steps}}=100$).

\textbf{Task and object.} Grasp-and-hold of a plastic cup of the same type used
in the demonstrations, placed empty at a marked position. The arm trajectory
delivers the hand; the policy performs the grasp and hold. The cup returns to
the mark between rollouts.

\subsection{Protocol}
\label{sec:protocol}
Sample size, metrics, thresholds, and the headline decision rule were fixed
before the first scored rollout, and all completed rollouts are reported.
The sample is $n=50$ rollouts, 25 per policy, in strict with/without
alternation with no extensions. Per-rollout success is defined as grasped and
held intact through the run, recorded by the operator as grasped, crushed,
dropped, or held, with video retained. The per-rollout grip-force metric is
the median, over hold-phase contact frames ($>0.3$\,N), of the mean
thumb--index force estimate in newtons, compared across policies by a
one-sided Mann--Whitney U test (with-force $<$ without-force, $\alpha=0.05$)
on successful rollouts. The pre-declared decision rule selects the headline:
if success differs (Fisher exact $p<0.05$, at least 25 percentage points), the
headline is reliability; if success saturates or does not differ but grip
force does, the headline is force regulation with success reported as no
difference. Both outcomes are reported regardless. Camera alignment and arm
pose were probed every eight rollouts, and exclusion rules for truncated logs
and stale frames were declared in advance.

\subsection{Baselines}
Three policies form the observation ladder. \textbf{Vision only}: the entire
state vector zeroed at training and deployment. \textbf{Vision + hand state}:
flex and IMU channels present, force zeroed (the without-force policy).
\textbf{Vision + hand state + fingertip force}: the full observation (the
with-force policy). All three share identical demonstrations, images, actions,
architecture, training steps, and seed.

\subsection{Results}
\label{sec:results}

\begin{figure}[t]
  \centering
  \includegraphics[width=\linewidth]{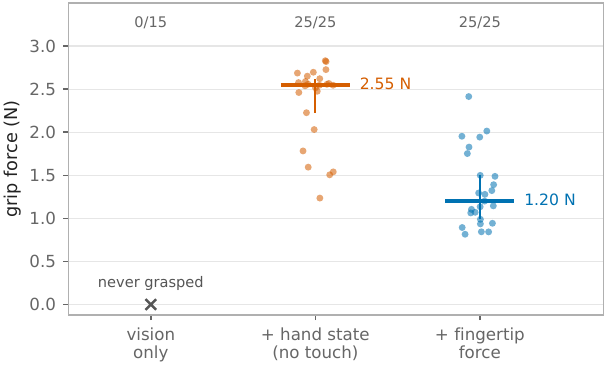}
  \caption{\textbf{Observation ablation on the cup grasp-and-hold task.} Tallies
  along the top are grasp success. The vision-only policy never acquires the
  grasp (0/15; interleaved controls in the same session succeeded 3/3), so it
  has no grip-force distribution. Adding hand-command state acquires the
  grasp, and adding the fingertip-force channel does not change whether the
  policy holds (25/25 for both) but changes how hard: median grip force falls
  from 2.55\,N to 1.20\,N ($-53\%$, one-sided Mann--Whitney $p<0.0001$). Each
  marker is one rollout ($n=25$ per policy); bars show median and interquartile
  range. Force is a current-based estimate calibrated against a reference scale
  (\S\ref{sec:force}). The vision-only arm was evaluated in a separate
  lamp-matched session with interleaved controls; force comparisons are
  same-session only.}
  \label{fig:chart}
\end{figure}

\begin{figure*}[t]
  \centering
  \includegraphics[width=\textwidth]{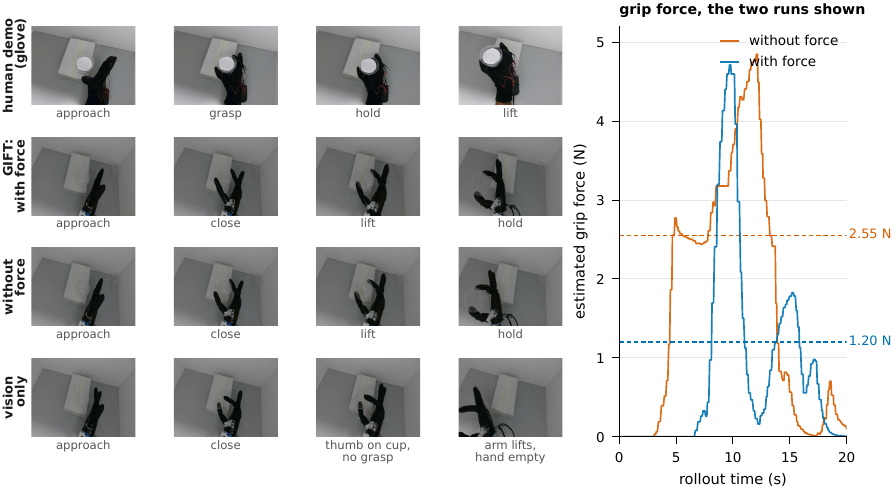}
  \caption{\textbf{Representative rollouts from the head camera.} Top row: a
  human demonstration with the sensing glove. Middle rows: the with-force and
  without-force policies on the robot hand, showing for each arm the run whose
  hold-phase median is closest to its arm's median (1.20\,N and 2.55\,N).
  Bottom row: the vision-only policy lands the thumb on the cup, bottoms out
  the close without a grasp, and the arm lifts an empty hand. Right: the logged
  grip-force estimate over the two state-policy runs shown, with their
  hold-phase medians dashed.}
  \label{fig:rollouts}
\end{figure*}

\begin{table}[t]
\centering
\caption{Observation ablation on the cup grasp-and-hold task. Hand-command
state acquires the grasp; the fingertip-force channel changes how hard the
policy holds, not whether it holds. Grip force is the median of per-rollout
hold-phase medians ($n=25$ per policy). All completed rollouts reported.}
\label{tab:ablation}
\begin{tabular}{lccc}
\toprule
Observations & Grasp & Hold & Grip force (N) \\
\midrule
Vision only              & 0/15  & ---   & --- \\
+ hand state (no touch)  & 25/25 & 25/25 & 2.55 \\
+ fingertip force        & 25/25 & 25/25 & \textbf{1.20} \\
\bottomrule
\end{tabular}
\end{table}


\textbf{Same success, half the grip force.}
All 50 rollouts completed with no exclusions. Task success saturated at
\textbf{25/25 for both policies}, so the decision rule selects force regulation
as the headline. The median of the per-rollout hold-phase grip-force
estimates was \textbf{1.20\,N} (IQR 0.99--1.50) with force inputs versus
\textbf{2.55\,N} (IQR 2.23--2.62) without, a \textbf{53\% reduction} (one-sided
Mann--Whitney U, $z=5.34$, $p<0.0001$; Fig.~\ref{fig:chart}; representative rollouts in
Fig.~\ref{fig:rollouts}). A drift check found no session trend
($\mathrm{corr}(\text{run},\text{median})=0.01$). Grasp-transient peaks were
statistically indistinguishable across policies and sat near the estimator's
ceiling, so they are not quoted as measurements. The without-force policy does
not fail the task; it succeeds by squeezing more than twice as hard as the
policy with force inputs, because nothing in its observation indicates when
the grip is sufficient. We do not compare either policy against the demonstrator's own
grip force: the glove FSRs and the deployment-side estimator are different
instruments, and only the policy-versus-policy comparison holds the instrument
fixed.

\textbf{Vision alone does not grasp.}
The vision-only policy failed to acquire the grasp in \textbf{0 of 15}
rollouts, while interleaved same-session control rollouts of the state-carrying
policies went \textbf{3/3} (Fisher exact $p\approx0.0012$; including two
mislabeled but otherwise valid runs, 0/16 versus 4/4, $p\approx0.0002$). The
failure signature was consistent (Fig.~\ref{fig:rollouts}, bottom row): the
thumb landed on the cup and the commanded close bottomed out without a grasp.
State-carrying policies grasped in every lighting condition tested across three
sessions; the vision-only policy failed in both sessions in which it was
tested. On this task the visual scene barely changes with hand configuration,
so hand-command state is what carries the timing of the grasp, and the
fingertip-force inputs are what regulate its strength
(Table~\ref{tab:ablation}).

\textbf{Mechanism.}
The action space is five finger positions; force enters the policy as an
observation and appears in no term of the loss. The 53\% reduction is therefore
force-informed imitation: observing force shifts which demonstrated positions
the policy reproduces. This locates the next step directly, since the
force-annotated demonstrations GIFT produces already support making force a
prediction target without new hardware.

\section{Limitations}
\label{sec:limits}

\textbf{Sensing hardware.} The glove's FSRs measure normal force only; shear,
which matters for slip and for in-hand reorientation, is not captured. FSRs
also exhibit drift and hysteresis, and the per-finger calibration must be
repeated when a sensor is replaced. On the robot side, the current-residual
estimator is a relative instrument: it is identical for both policies, which
is what the comparison requires, but absolute newton values are approximate,
and the estimate saturates near 5\,N, censoring the grasp transient. A load-cell
reference object shared by glove and robot would put both instruments on one
scale and enable direct human-to-robot force comparison.

\textbf{Task scope.} All results are on one object at one position with a
fixed arm trajectory; the policy controls the hand only. Success saturated at
25/25 for both policies, so this evaluation measures force regulation, not
reliability. Success was recorded by a single operator from a fixed
alternation order, with video retained; the evaluation is not blinded.
Extending to object variation, arm control, and tasks where over-gripping
causes visible failure (crushing, slipping) is the natural next evaluation.

\section{Conclusion}
We presented GIFT, a pipeline for human-to-robot skill transfer in which the
interface between human and robot is a physical unit: calibrated fingertip
force captured on a human hand with a wearable sensing glove is transferred to
a robot hand with no tactile sensors, using actuator-current residuals as the
deployment-side force estimate. On a cup grasp-and-hold task, two policies
trained on the same demonstrations reach the same success, and the policy with
force inputs holds at half the estimated grip force. The observation ablation
decomposes the channels cleanly: vision alone does not grasp, hand-command
state acquires the grasp, and the fingertip-force inputs turn a successful
squeeze into a controlled hold. Because the policy uses a glove-space state
representation, force is captured as a physical unit, and the deployment-side
estimate comes from a signal every position-controlled hand provides, the
pipeline carries no dependency on the sensing or actuation hardware used
here.

\bibliographystyle{IEEEtranN}
\bibliography{refs}

\end{document}